\documentclass{sig-alternate}

\conferenceinfo{GECCO'07,}{July 7--11, 2007, London, England, United Kingdom.}
\CopyrightYear{2007}
\crdata{978-1-59593-697-4/07/0007}

\usepackage{tabularx}
\usepackage{graphicx}
\usepackage{amsmath}
\usepackage{url}

\graphicspath{{matlab/}}

\DeclareMathOperator*{\argmax}{argmax}

\def\RR{{\rm I\hspace{-0.50ex}R}}
\def\Off{{\em Off}-EEL}
\def\On{{\em On}-EEL}

\begin{document}

\title{Ensemble Learning for Free with Evolutionary Algorithms ?}

\numberofauthors{2} 

\author{
\alignauthor
Christian Gagn\'e\titlenote{This work has been mainly realized during a postdoctoral fellowship of Christian Gagn\'e at the University of Lausanne.}\\
  \affaddr{Informatique WGZ Inc.,}\\
  \affaddr{819 avenue Monk,}\\
  \affaddr{Qu\'ebec (QC), G1S 3M9, Canada.}\\
  \email{christian.gagne@wgz.ca}
\alignauthor
Mich\`ele Sebag\\
  \affaddr{\'Equipe TAO -- CNRS UMR 8623 / INRIA Futurs,}\\
  \affaddr{LRI, Bat. 490, Universit\'e Paris Sud,}\\
  \affaddr{F-91405 Orsay Cedex, France.}\\
  \email{michele.sebag@lri.fr}
\and
\alignauthor
Marc Schoenauer\\
  \affaddr{\'Equipe TAO -- INRIA Futurs / CNRS UMR 8623,}\\
  \affaddr{LRI, Bat. 490, Universit\'e Paris Sud,}\\
  \affaddr{F-91405 Orsay Cedex, France.}\\
  \email{marc.schoenauer@lri.fr}
\alignauthor
Marco Tomassini\\
  \affaddr{Information Systems Institute,}\\
  \affaddr{Universit\'e de Lausanne,}\\
  \affaddr{CH-1015 Dorigny, Switzerland.}\\
  \email{marco.tomassini@unil.ch}
}

\date{11th April 2007}

\maketitle

\begin{abstract}
Evolutionary Learning proceeds by evolving a population 
of classifiers, from which it generally returns (with some notable exceptions) 
the single best-of-run classifier as final result. In the meanwhile, 
Ensemble Learning, one of the most efficient approaches 
in supervised Machine Learning for the last decade, 
proceeds by building a population of diverse classifiers.
Ensemble Learning with Evolutionary Computation thus receives 
increasing attention. The Evolutionary Ensemble Lear\-ning (EEL) approach
presented in this paper features two contributions. 
First, a new fitness function, inspired by co-evo\-lu\-tion
and enforcing the classifier diversity, is presented. Further, a new selection
criterion based on the classification margin is proposed. This criterion is
used to extract the classifier ensemble from the final population only (\Off)
or incrementally along evolution (\On).
Experiments on a set of benchmark problems show that \Off\ outperforms 
single-hypothesis evolutionary learning and state-of-art 
Boosting and generates smaller classifier ensembles.
\end{abstract}


\section{Introduction}

Ensemble Learning, one of the main advances in Supervised Machine Learning since
the early 90's, relies on: i) a weak learner (extracting hypotheses, aka 
classifiers, with error probability less than $1/2 - \epsilon$, $\epsilon > 0$);
ii) a diversification heuristics used to extract sufficiently diverse classifiers; 
iii) a voting mechanism, aggregating the diverse classifiers constructed \cite{Breiman1998,Freund1996}.
%% Je mets partout classifiers pour eviter la confusion.
If the classifiers are sufficiently diverse and their errors are 
independent, then their majority vote %of the classifiers 
will reach  
an arbitrarily low error rate on the training set 
as the number of classifiers increases \cite{Esposito2003}. Therefore, up
to some restrictions on the classifier space \cite{Vapnik1998}, 
the generalization error will also be low\footnote{In practice, the generalization error is estimated from the error on a test set, disjoint from the 
training set. The reader is referred to \cite{Dietterich1998} for a comprehensive
discussion about the comparative evaluation of learning algorithms.}.

The most innovative aspect of Ensemble Learning w.r.t. the Machine Learning 
literature concerns the diversity requirement, implemented 
through parallel or sequential heu\-ris\-tics. In Bagging, diversity is 
enforced by considering independent sub-samples of the training set, 
and/or using different learning parameters \cite{Breiman1998}.  
Boosting iteratively constructs a sequence of
classifiers, where each classifier focuses on the examples misclassified by the 
previous ones \cite{Freund1996}.

Diversity is also a key feature of Evolutionary Computation (EC): in contrast with
all other stochastic optimization approaches, evolutionary algorithms 
proceed by evolving a population of solutions, and the 
diversity thereof has been stressed as a key factor of success  since the 
beginnings of EC. Deep similarities between
Ensemble Learning and EC thus appear; in both 
cases, diversity is used to escape from local minima, where any single ``best''
solution is only too easily trapped. Despite this similarity, Evolutionary
Learning has most often (with some notable exceptions, see 
\cite{Holland1986,Iba1999,Liu2000} among others) focused 
on single-hypothesis learning, where some single best-of-run hypothesis is 
returned as the solution.

However, the evolutionary population itself could be used as a pool
for recruiting the elements of an ensemble, enabling ``Ensemble Learning
for Free''. Previous work along this line will be described in Section 
\ref{sec:RelatedWork}, mostly based on using an evolutionary algorithm 
as weak learner \cite{Keijzer2000}, or using evolutionary diversity-enforcing 
heuristics \cite{Iba1999,Liu2000}.

In this paper, the ``Evolutionary Ensemble Learning For Free'' claim 
is empirically examined along two directions. The first direction is that 
of the classifier diversity; a new learning-oriented fitness 
function is proposed, inspired by the co-evolution framework \cite{Hillis1990}
and generalizing the di\-ver\-si\-ty-en\-for\-cing fitness proposed by \cite{Liu2000}.
The second direction is that of the selection of the ensemble classifiers 
within the evolutionary population(s).
%It has been shown that a learning 
%ensemble can be pruned, particularly in the Boosting case, with no impact on the 
%accuracy of the ensemble \cite{Dietterich2000}. 
Selecting the best classifiers in a pool amounts to a feature selection 
problem, that is, a %NP-hard 
combinatorial optimization problem \cite{Guyon2006}. 
A greedy set-covering approach is used, build on a margin-based 
criterion inspired by Schapire \emph{et al.} \cite{Schapire1998}. 
Finally, the paper presents two Evolutionary Ensemble Learning (EEL)
approaches, called \Off\ and \On, respectively tackling the selection 
of the ensemble classifiers in the final population, 
or along evolution.

Paper structure is as follows. Section \ref{sec:RelatedWork} reviews 
and discusses some work relevant to Evolutionary Ensemble Learning. 
Section \ref{sec:MethodsProposed} describes the two proposed approaches \Off\ and \On, 
introducing the specific fitness function and the ensemble classifier 
selection procedure. Experimental results based on benchmark problems from 
the UCI repository
%, using separating hyperplanes as classifier search space, 
are reported in Section \ref{sec:Experiments}. 
%\Off\ and \On\ are compared 
%with standard evolutionary learning and standard Boosting, using decision
%stumps as classifier search space (see \cite{Schapire1998}). While Boosting
%and EEL approaches obtain similar generalization errors, 
%EEL does so with significantly less classifiers in the ensemble. 
The paper 
concludes with some perspectives for further research, discussing the 
priorities for a tight coupling of Ensemble Learning with Evolutionary 
Optimization in terms of dynamic systems \cite{Rudin2004}.

\section{Related Work}
\label{sec:RelatedWork}

Interestingly, some early approaches in Evolutionary Learning were 
rooted on Ensemble Learning ideas\footnote{
Learning Classifier Systems (LCS, \cite{Holland1986,Holmes2002}) are mostly devoted to Reinforcement Learning, as 
opposed to Supervised Machine Learning; therefore they
will not be considered in the paper.}.
The Michigan approach \cite{Holland1986} evolves a population
made of rules, whereas the Pittsburgh approach evolves a population made of
sets of rules. What is gained in flexibility and tractability in the 
Michigan approach is compensated by the difficulty of assessing a 
single rule, for the following reason. A rule usually only covers a part of 
the example space; gathering the best rules (e.g. the rules with highest accuracy)
does not result in the best ruleset. Designing
an efficient fitness function, such that a good quality ruleset could
be extracted from the final population, was found a tricky task.

In the last decade, Ensemble Learning has been explored within
Evolutionary Learning, chiefly in the context of Genetic Programming (GP).
A first trend directly inspired from Bagging and Boosting aims at
reducing the fitness computation cost \cite{Folino2003,Iba1999}
and/or dealing with datasets which do not fit in memory \cite{Song2005}.
For instance, Iba \cite{Iba1999} divided
the GP population into several sub-populations which are evaluated on 
subsets of the training set. Folino \emph{et al.} \cite{Folino2003} likewise 
sampled the training set in a Bagging-like mode in the context of 
parallel cellular GP. Song \emph{et al.} \cite{Song2005} used Boosting-like
heuristics to deal with training sets that do not fit in memory; the training 
set is divided into folds, one of which is loaded in memory and periodically replaced; 
at each generation, small subsets are selected from the current fold 
to compute the fitness function, where 
the selection is nicely based on a mixture of 
uniform and Boosting-like distributions. 

The use of Evolutionary Algorithms as weak learners with\-in a standard Bagging
or Boosting approach has also been investigated. Boosting approaches for GP have been applied for instance to classification \cite{Paris2001} or symbolic regression \cite{Keijzer2000}: each run delivers a GP tree 
minimizing the weighted sum of the training errors, and the weights 
were computed as in standard Boosting \cite{Freund1996}. While such ensembles of 
GP trees result, as expected, in a much lower variance of the performance, 
they do not fully exploit the population-based nature of GP, as 
independent runs are launched to learn successive classifiers.

% As already mentioned, the cornerstone of Ensemble Learning is the diversity 
% of the classifiers in the ensemble \cite{Dietterich2000,Brown2005}. 
% Kuncheva and Whitaker \cite{Kuncheva2003} have identified some ten diversity measures, including four
% pairwise measures and six non-pairwise measures. Although these measures are 
% indeed relevant to study the output of an ensemble learning algorithm, they
% still present some limitations in order to successfully evolve such an
% ensemble. On one hand, non-pairwise measures assess the diversity of 
% the ensemble; they deliver little information as to the contribution of 
% a particular classifier to the ensemble, and therefore cannot be used directly 
% as fitness function in a Michigan-style evolutionary learning approach.
% On the other hand, pairwise measures can be viewed
% as phenotypic distances between two classifiers, where the 
% classifier phenotype is the vector of its values on the training set.
% While pairwise distances can be exploited within a fitness sharing 
% framework (see below), they can only emulate Bagging approaches since 
% all training examples are considered equally important. This contrasts with
% the spirit of Boosting, where the classifier diversity is obtained by 
% putting the stress on those examples which are currently misclassified.

Liu \emph{et al.} \cite{Liu2000} proposed a tight coupling between Evolutionary
Algorithms and Ensemble Learning. They constructed an ensemble of Neural 
Networks, using a modified back-propagation algorithm 
to enforce the diversity of the networks; specifically, the back-propagation
aims at both minimizing the training error and maximizing the negative correlation of
the current network with respect to the current population. Further, the 
fitness associated to each network is the sum of the weights of all examples it 
correctly classifies, where the weight of each example is inversely proportional
to the number of classifiers that correctly classify this example. While
this approach
nicely suggests that ensemble learning is a Multiple Objective Optimization (MOO) 
problem (minimize the error rate and maximize the diversity), it classically
handles the MOO problem as a fixed weighted sum of the objectives. 

The MOO perspective was further investigated by Chandra and Yao 
in the DIVACE system, 
a highly sophisticated system for the multi-level evolution 
of ensemble of classifiers \cite{Chandra2006b,Chandra2006a}. 
In \cite{Chandra2006a}, the top-level evolution 
simultaneously minimizes the error rate (accuracy) and maximizes the negative correlation
(diversity). In \cite{Chandra2006b}, the negative correlation-inspired criterion is replaced 
by a \emph{pairwise failure crediting}; the difference concerns the misclassification of examples that are 
correctly classified by other classifiers.
Finally, the ensemble is constructed either by keeping all classifiers in the 
final population, or by clustering the final population (after their phenotypic
distance) and selecting a classifier in each cluster.

While the MOO perspective nicely captures the interplay of the 
accuracy and diversity goals within Ensemble Learning,
 the selection of the classifiers in the 
genetic pool as done in \cite{Chandra2006b,Chandra2006a} does not fully exploit the 
possibilities of evolutionary optimization, in two respects. On the one hand,
it only considers the final population that usually involves up to a few hundred classifiers, while learning ensembles commonly involve some thousand classifiers.
On the other hand, clustering-based selection proceeds on the basis of the 
phenotypic distance between classifiers, considering again that all examples
are equally important, while the higher stress put on harder examples is 
considered the source of the better Boosting efficiency \cite{Dietterich2000}.
%Accordingly, two ensemble selection procedures will be defined in the next section.

\section{Ensemble Learning for Free}
\label{sec:MethodsProposed}

After the above discussion, Evolutionary Ensemble Learning (EEL) involves
two critical issues: i) how to enforce both the predictive accuracy and 
the diversity of the classifiers in the population, and across generations; 
ii) how to best select the ensemble classifiers, 
from either the final population only or all along evolution. 

Two EEL frameworks have been designed to study these interdependent issues. The first one dubbed \emph{Offline Evolutionary Ensemble Learning} (\Off) constructs the ensemble 
from the final population only. The second one, called \emph{Online Evolutionary Ensemble Learning} (\On), gradually constructs the classifier ensemble as a selective archive of evolution, where some classifiers are added to the archive at each generation.
 
Both approaches combine a standard generational evolutionary algorithm with two
interdependent components: a new  diversity-enhancing 
fitness function, and a selection mechanism. The fitness function, presented in 
Section \ref{sec:Diversity} and generalizing the fitness devised by Liu \emph{et al.}
\cite{Liu2000}, is inspired from co-evolution \cite{Hillis1990}. The selection
process is used to extract a set of classifiers from either the final population (\Off) or the current archive plus the current population (\On), and proceeds by 
greedily maximizing the ensemble margin (Section \ref{sec:Criterion}).

Only binary or multi-class classification problems are considered in this 
paper. The decision of the classifier ensemble is the majority vote among the 
classifiers (ties being arbitrarily broken).

\subsection{Diversity-enforcing Fitness}
\label{sec:Diversity}

Traditionally, Evolutionary Learning maximizes the number of correctly 
classified training examples (or equivalently minimizes the error rate).
However, examples are not equally informative;
% \footnote{The emerging 
% field of Active Learning \cite{ActiveLearning} is specifically
% concerned with the search for the most informative examples.}; 
% Cette footnote pourrait disparaitre selon moi - MARC (Michèle OK)
therefore a rule correctly 
classifying a hard example (e.g. close to the frontiers of the target 
concept) is more interesting and should be more rewarded
%, everything else being equal, 
than a 
rule correctly classifying an example which is correctly classified 
by almost all rules. 

Co-evolutionary learning, first pioneered by Hillis \cite{Hillis1990}, 
nicely takes advantage of the above remark, gradually forging 
more and more difficult examples to enforce the discovery of high-quality
solutions. Boosting proceeds along the same lines, gradually 
putting the stress on the examples which have not been successfully
predicted so far.

A main difference between both frameworks is that
Boosting exploits a finite set of labelled examples, while
co-evol\-utionary learning has an infinite supply of labelled examples
(since it embeds the oracle). A second difference is that the difficulty
of an example depends on the whole sequence of classifiers in 
Boosting, whereas it only
depends on the current classifier population in co-evolution. In other words, 
Boosting is a memory-based process, while co-evolutionary learning 
is a memoryless one. Both approaches thus suffer from opposite weaknesses. 
Being a memory-based process, Boosting can be misled 
by noisy examples; consistently misclassified, these examples eventually get 
heavy weights and thus destabilize the Boosting learning process. Quite the 
contrary, co-evolution can forget what has been learned during early stages
and specific heuristics, e.g. the so-called Hall-of-Fame, archive of best-so-far individuals, are required to 
prevent co-evolution from cycling in the learning landscape \cite{Paredis1997}.

Based on these ideas, the fitness of classifiers is defined in this work from
a set of reference classifiers noted $\cal Q$.
% (defined in Sections \ref{Off} and \ref{On}). 
The hardness of every training example $x$ is 
measured after the number of classifiers in $\cal Q$ which 
misclassify $x$. The fitness of every classifier $h$ is then measured
by the cumulated hardness of the examples that are correctly classified by $h$.

Three remarks can be made concerning this fitness function. Firstly, 
contrasting with standard co-evolution, there is no way classifiers can 
``unlearn'' to classify the training examples, since the training set is fixed.
Secondly, as in Boosting, the fitness of a classifier reflects its diversity with respect to
the reference set. 
Lastly, the classifier fitness function is highly multi-modal compared to the 
simple error rate: good classifiers might correctly classify many easy 
examples, or sufficiently many hard enough examples, or a few very hard examples.

Formally, let ${\cal E}=\{({\bf x}_i,y_i),~{\bf x}_i\in {\cal X},~y_i\in Y,~i=1\ldots n\}$ denote the training set (referred to as set of fitness cases in the GP context); each fitness case or example $({\bf x}_i,y_i)$ is composed of an instance ${\bf x}_i$ belonging to the instance space ${\cal X}$ and the associated label $y_i$ belonging to a finite set $Y$. Any classifier $h$  
is a function mapping the instance space $\cal X$ onto $Y$. The  
loss function $\ell$ is defined as $\ell : Y \times Y \mapsto \RR$, where $\ell (y,y')$ is the (real valued) error cost of predicting label $y$ instead of the true label $y'$. 

The hardness or weight of every training example $({\bf x}_i,y_i)$, noted $w^{\cal Q}_i$, or $w_i$ when the reference set $\cal Q$ is clear from the context, is the average loss incurred by the reference classifiers on $({\bf x}_i,y_i)$: 
\begin{equation}
w_i = \frac{1}{|{\cal Q}|} \sum_{h \in {\cal Q}} \ell(h({\bf x}_i),y_i).
\end{equation}

The cumulated hardness fitness $\cal F$ is finally defined as follows: ${\cal F}(h)$ is the sum over all training examples that are correctly classified by $h$, of their weight $w_i$ raised to power $\gamma$. Parameter $\gamma$ governs the importance of the weights $w_i$ (the cumulated hardness boils down to the number of correctly classified examples for $\gamma=0$) and thus the diversity pressure.  
\begin{equation}
{\cal F}(h) = \sum_{i = 1\ldots n \atop h({\bf x}_i) = y_i} w_i^\gamma 
\label{equ:DiversityFitness}
\end{equation}

Parameter $\gamma$ can also be adjusted depending on the level of noise in the dataset. As noisy examples typically reach high weights, increasing the value of $\gamma$ might lead to retain spurious hypotheses, which happen to correctly classify a few noisy examples. When $\ell$ is set to the step loss function ($\ell(y,y')=0$ if $y=y'$, $1$ otherwise) and $\gamma$ is set to 1, the above fitness function is the same as the one used by Liu \emph{et al.} \cite{Liu2000}. The value of $\gamma$ 
is set to $2$ in the experiments (Section \ref{sec:Experiments}).

\subsection{Ensemble Selection}
\label{sec:Criterion}

As noted earlier on, the selection of classifiers in a pool ${\cal H}=\{h_1,\ldots,h_T\}$ in order to form an efficient ensemble is formally equivalent to a feature selection problem. The equivalence is seen by replacing the initial instance space $\cal X$ with the one defined from the classifier pool, where each instance ${\bf x}_i$ is redescribed as the vector $(h_1({\bf x}_i),\ldots,h_T({\bf x}_i))$. Feature selection algorithms \cite{Guyon2006} could thus be used for ensemble selection; unfortunately, feature selection is one of the most difficult Machine Learning problems.

Therefore, a simple greedy selection process is used in this paper to select the classifiers in the diverse pools considered by the \Off\ (Section  \ref{sec:OffEEL}) and \On\ (Section \ref{sec:OnEEL}) algorithms. The novelty is the selection
criterion, generalizing the notion of margin \cite{Gilad-Bachrach2004,Schapire1998} to an ensemble of examples as follows.
\begin{figure*}[t]
\caption{Pseudo-code of \texttt{Ensemble-Selection}(Classifier pool $\cal H$, training set ${\cal E}$, initial classifier ensemble ${\cal L}_0$).}
\label{fig:EnsembleConstrAlgo}
\begin{center}\fbox{\parbox{\linewidth}{
\begin{enumerate}
\item Let $t=1$, and ${\cal H}_1$ be the set $\cal H$ with duplicate individuals removed
\item While ${\cal H}_t$ is not empty:
\begin{enumerate}
\item Let $h^*_t = \argmax_{h\in {\cal H}_t}\left({\cal L}_{t-1} \cup \{h\} \right)$ after the margin-based order relation of Equation \ref{equ:Selectionmax}
\item Let ${\cal H}_{t+1} = {\cal H}_t \backslash \{h^*_t\}$ (remove $h^*_t$ from ${\cal H}_{t}$)
\item Let ${\cal L}_t = {\cal L}_{t-1} \cup \{ h^*_t \}$ (and add it to ${\cal L}_{t}$)
\item $t=t+1$
\end{enumerate}
\item Return ${\cal L}^*$, the classifier ensemble in $\{{\cal L}_{0}\ldots{\cal L}_{t-1}\}$ that achieves the lowest error rate on ${\cal E}$, selecting the smallest ensemble in case of ties.
\end{enumerate}
}}
\end{center}
\label{fig:Selection}
\end{figure*}

Formally, let $\cal L$ denote the current ensemble, initialized to the classifier $h^*$ with minimum error rate in $\cal H$. For each example $({\bf x}_i,y_i)$, let its margin $m_i$ be defined as follows. Let $y'_i$ be the class most frequently associated to ${\bf x}_i$ by the classifiers in $\cal L$, such that $y'_i$ is different from the true class $y_i$. Let $c_i$ (respectively $c'_i$) denote the number of classifiers in $\cal L$ associating class $y_i$ (resp. $y'_i$) to ${\bf x}_i$. Then margin $m_i$ is defined as $c_i - c'_i$. A positive margin thus denotes the fact that the example is correctly classified by the majority vote; the higher the margin, the more confident the ensemble prediction. Conversely, a negative margin denotes an error; the ensemble misclassifies the example as belonging to class $y'_i$; the more negative the margin, the more classifiers need to be added to the ensemble in order to correctly classify ${\bf x}_i$.

Let $K$ denote the number of classes of the problem and $|A|$ the size of a set $A$. The above definitions then read:
\begin{eqnarray}
y'_i & = & \argmax_{k=1\ldots K \atop k\neq y_i} \left|\{h_j({\bf x}_i) = k,~ h_j \in {\cal L}\}\right|,\\
m_i  & = & |\{h_j({\bf x}_i) = y_i,~h_j \in {\cal L}\}| - \nonumber \\
     &   & \hspace{1in} |\{h_j({\bf x}_i) =  y'_i ,~ h_j \in {\cal L}\}|.
\end{eqnarray}
Initially, the quality of ensemble $\cal L$ was measured after its minimum 
margin when $({\bf x}_i,y_i)$ ranges over the training set, and the selection
process aimed at maximizing the minimum margin likewise 
Boosting \cite{Rudin2004}. However, it turned out experimentally 
that the minimum margin alone is too coarse a criterion, leading to many ties. 
Thus, a finer grained criterion, based on the margin histogram, has finally
been defined.
 
Let $c({\cal L},m)$ denote the number of training examples with margin $m$ 
after $\cal L$. An order relation on classifier ensembles ${\cal L}$ and ${\cal L}'$ 
can then be defined by comparing  $c({\cal L},m)$ and $c({\cal L}',m)$ for 
increasing values of $m$; the best ensemble is the one with lesser number of 
examples with the smallest margin. % In case of equality, the best ensemble is the smallest one.
\begin{equation}
% \nonumber
% {\cal L} < {\cal L}' ~\mbox{iff one of the following two conditions holds}\\
%\hspace*{1in}
% \label{equ:Selectionmax} (i)~ \forall m, c({\cal L},m) = c({\cal L}',m) \mbox{~and~} |{\cal L}| > |{\cal L}'|\\
%\hspace*{1in}
%\label{equ:Selectionegal}
% (ii)~ \exists~ m_0~ \mbox{s.t.}~ \left\{ \begin{array}{l}
% \forall m < m_0, ~ c({\cal L},m) = c({\cal L}',m)\\ c({\cal L},m_0) > c({\cal L}',m_0)\end{array}\right.
{\cal L} < {\cal L}'~\mbox{iff}~\exists~ m_0~ \mbox{s.t.}~ \left\{ \begin{array}{l}\forall m < m_0, ~ c({\cal L},m) = c({\cal L}',m)\\ c({\cal L},m_0) > c({\cal L}',m_0)\end{array}\right. \label{equ:Selectionmax}
\end{equation}

% The selection process iteratively selects in $\cal H$ the classifier $h^*$ 
% such that it leads to a maximal classifier ensemble after the above order 
% relation. 

The pseudo-code of the ensemble selection algorithm is displayed in 
Figure \ref{fig:EnsembleConstrAlgo}. It starts with a classifier pool ${\cal H}$, a set of training examples ${\cal E}$ and an initial set of classifiers ${\cal L}_0$. It then iteratively moves all classifiers from ${\cal H}$ into ${\cal L}$, based on the above order on ensembles. Ultimately, the ensemble with lowest error rate on ${\cal E}$ in the  ensemble sequence ${\cal L}_0\ldots {\cal L}_{t-1}$ is selected.

\subsection{Offline Evolutionary Ensemble Learning}
\label{sec:OffEEL}

\Off\ is a two-step process. It firstly runs a standard evolutionary learning algorithm. The approach does not make any requirement on the genetic search space, that is the classifier space; the designer can run \Off\ on the top of her favorite evolutionary learning algorithm, searching for linear classifiers, neural nets, rule systems, or genetic programs. 
%In order to facilitate the analysis of the results, experiments will consider the straightforward space of linear classifiers.
The only required modification concerns the fitness function, which is set to the diversity-enhancing fitness described in Section \ref{sec:Diversity}, taking the whole current population as set of reference classifiers. 
%Another interpretation for this fitness function when parameter $\gamma$ is set to $1$\footnote{Up to normalization as singularities must be avoided when some examples are misclassified by every reference classifiers.}, is that every example has the same credit, which is equally divided among the classifiers which correctly classify it. 
In contrast with Boosting, the process does not maintain any memory about the examples; their weights are recomputed from scratch at each generation. While Boosting  might result in exponentially increasing the  weight of hard or possibly noisy examples, \Off\ thus keeps the weight of each training example bounded, and thereby avoids the instability due to the data noise.

The second step achieves the ensemble selection based on the margin-based criterion (Section \ref{sec:Criterion} and Figure \ref{fig:EnsembleConstrAlgo}).
It uses the final population as pool of classifiers $\cal H$, and initializes
the classifier ensemble to the classifier $h^*$ that has the  
smallest error rate on the training set in the population (${\cal L}_0 = \{ h^*\}$).

\subsection{Online Evolutionary Ensemble Learning}
\label{sec:OnEEL}

In contrast with \Off, \On\ interleaves evolutionary learning and ensemble 
selection; at each generation the classifier ensemble is updated using the current population.

At generation 1, the classifier ensemble is initialized to the classifier that  minimizes the error rate on the training set. In further generations, 
the current population is evolved using the diversity-enhancing fitness function with the current ensemble as reference set (Section \ref{sec:Diversity}), and the ensemble selection algorithm (Figure \ref{fig:EnsembleConstrAlgo}) is launched, 
using the current population as classifier pool $\cal H$, and 
the current classifier ensemble as ${\cal L}_0$.
%
% Mais alors, des que l'ensemble classifie bien tout le monde, on n'ajoute
% lus personne ???
%
% CG : effectivement, bon point, lorsque le error rate sur le training set est 0, on n'ajoute
% plus personne. Mais en pratique ce n'est pas survenu dans nos expériences.
%
The pseudo-code of \On\ is given in Figure \ref{fig:HighLevelOnEEL}.

\begin{figure}[t]
\caption{Pseudo-code of \texttt{On-EEL}(training set ${\cal E}$).}
\label{fig:HighLevelOnEEL}
\begin{center}
\fbox{\parbox{\linewidth}{
\begin{enumerate}
\item Let ${\cal P}_1$ be the first evolutionary population, and $h^*$ the classifier with minimal error rate on $\cal E$.
\item ${\cal L}_1$ = \texttt{Ensemble-Selection}(${\cal P}_{1}, {\cal E}, \{ h^*\}$)
\item For $t = 2 \ldots T$:
\begin{enumerate}
\item Evolve ${\cal P}_{t-1} \rightarrow {\cal P}_{t}$, using ${\cal L}_{t-1}$ as reference set. 
\item ${\cal L}_t$ = \texttt{Ensemble-Selection}(${\cal P}_{t}, {\cal E}, {\cal L}_{t-1}$)
\end{enumerate}
\item Return ${\cal L}_T$.
\end{enumerate}
}}
\end{center}
\end{figure}

%
% Ceci saute : il n'y a plus qu'une seule valeur de \gamma, n'est-ce pas ?
% CG: Oui, c'est bien ca!
%
%Interestingly, a higher value of parameter $\gamma$ (Section \ref{sec:Diversity}) was found beneficial compared to the \Off\ setting. This fact can be explained from the exploration versus exploitation trade-off. The ensemble selection algorithm is performed every generation (as opposed to once in \Off), thus increasing the exploitation bias. For the sake of exploration, the diversity in the pool used for the ensemble selection, that is, the previous archive plus the current population, should thus be favored. Indeed, the current population is evolved to maximize its diversity compared to the reference set. But the reference set is the previous archive in \On, that is a restricted set of high quality classifiers, whereas it is the whole current population in \Off. A higher pressure toward diversity through a higher $\gamma$ value thus allows \On\ to resist the lesser diversity of the reference set. A few preliminary experiments led to $\gamma=2$.

Notably, \Off\ and \On\ achieve different Exploration {\em vs} Exploitation 
trade-offs. In \Off, the set of reference classifiers is the current 
population; 
the fitness function thus favors both accurate and diverse classifiers in 
each generation. The ensemble selection algorithm is launched only once, 
on a high quality and diversified pool of classifiers. 

In \On, the set of reference classifiers is the current classifier
ensemble; like in Boosting, the goal is to find classifiers which overcome the
errors of the past classifiers. While the ensemble selection algorithm 
is launched at every generation, it uses the 
biased current population as classifier pool. 
In fact, \On\ addresses a dynamic optimization 
problem; if the classifier ensemble significantly changes between one generation
and the next, the fitness landscape will change accordingly and several
evolutionary generations might be needed to accommodate this change. On the
other hand, as long as the current population does not perform well, the 
ensemble selection algorithm is unlikely to select further classifiers
in the current ensemble; the fitness landscape thus remains stable. 
The population diversity does not
directly result from the fitness function as in the \Off\ case; rather, 
it relates with the dynamic aspects of the fitness function. 

\section{Experimental setting}
\label{sec:Experiments}

This section describes the experimental setting used to assess the EEL framework.

\subsection{Datasets}

Experiments are conducted on the six UCI datasets \cite{Newman1998} presented in Table \ref{tab:DataSetsDescription}.
\begin{table*}[htbp]
\caption{UCI datasets used for the experimentations.}
\label{tab:DataSetsDescription}
\begin{center}
\begin{tabularx}{\linewidth}{ccccX}
          &         & \#       & \#      &\\
Dataset   & Size    & features & classes & Application domain\\\hline
{\tt bcw} & $683$   & $9$      & $2$     & Wisconsin's breast cancer, $65\:\%$ benign and $35\:\%$ malignant.\\
{\tt bld} & $345$   & $6$      & $2$     & BUPA liver disorders, $58\:\%$ with disorders and $42\:\%$ without disorder.\\
{\tt bos} & $508$   & $13$     & $3$     & Boston housing, $34\:\%$ with median value $v<18.77$ K\$, $33\:\%$ with $v\in]18.77,23.74]$, and $33\:\%$ with $v>23.74$.\\
{\tt cmc} & $1473$  & $9$      & $3$     & Contraceptive method choice, $43\:\%$ not using contraception, $35\:\%$ using short-term contraception, and $23\:\%$ using long-term contraception.\\
{\tt pid} & $768$   & $8$      & $2$     & Pima indians diabetes, $65\:\%$ tested negative and $35\:\%$ tested positive for diabetes.\\
{\tt spa} & $4601$  & $57$     & $2$     & Junk e-mail classification, $61\:\%$ tested non-junk and $39\:\%$ tested junk.\\
\end{tabularx}
\end{center}
\end{table*}
The performance of each algorithm is measured after a standard stratified 10-fold cross-validation procedure. The dataset is partitioned into 10 folds with same class distribution. Iteratively, all folds but the $i$-th one are used to train a classifier, and the error rate of this classifier on the remaining $i$-th fold is recorded. The performance of the algorithm is averaged over 10 runs for each fold, and over the 10 folds.

\subsection{Classifier Search Space}

As mentioned earlier on, evolutionary ensemble learning can accommodate any type of classifier; \Off\ and \On\ could consider neural nets, genetic programs or decision lists as genotypic search space. Our experiments will consider the most straightforward classifiers, namely separating hyperplanes, as these can easily be inspected and compared. Formally, let ${\cal X} = \RR^d$ be the instance space, a separating hyperplane classifier $h$ is characterized as $({\bf w},b) \in \RR^d \times \RR$ with $h({\bf x})={<{\bf w},{\bf x}>}-{b}$~ ($<{\bf w},{\bf x}>$ denotes the scalar product of ${\bf w}$ and ${\bf x}$). The search for a separating hyperplane is amenable to quadratic optimization, with:
\begin{equation}
{\cal F}(h) = \sum_{i = 1\ldots n} (h({\bf x}_i) - y_i)^2.\label{eq:convex}
\end{equation}
As the above optimization problem can be tackled using standard optimization algorithms, it provides a well-founded baseline for comparison.
% \footnote{Note that this formalization is over-constrained. Actually, one would be satisfied with $\forall i = 1\ldots n, ~h({\bf x}_i)\times y_i > 0$, that is, $h({\bf x}_i)$ is positive iff the example is positive ($y_i = 1$), and $h({\bf x}_i)$ is negative otherwise ($y_i = -1)$. However, the set of linear inequations defined by the training examples does not necessarily admit a solution (e.g. if there is some noise in the training examples). The quadratic formalization was thus preferred.}
Specifically, the first goal of the experiments is thus to assess the merits of evolutionary ensemble learning against three other approaches.

The first baseline algorithm referred to as Least Mean Square (LMS) uses a stochastic gradient algorithm to determine the optimal separating hyperplane in the sense of criterion given by Equation \ref{eq:convex} (see pseudo-code in Figure \ref{fig:AlgoLMS}).

The second baseline algorithm is an elementary evolutionary algorithm, producing the best-of-run separating hyperplane such that it minimizes the (training) error rate\footnote{For 3-classes problems, e.g. {\tt bos} or {\tt cmc}, the classifier is characterized as two hyperplanes, respectively separating class 0 (resp. class 1) from the other two classes. In case of conflict (the example is simultaneously classified in class 0 by the first classifier and in class 1 by the second classifier), the tie is broken arbitrarily.}.

The third reference algorithm is the prototypical ensemble learning algorithm, namely AdaBoost with its default parameters \cite{Freund1996}. AdaBoost uses simple decision stumps \cite{Schapire1998} baseline algorithm as weak learner (more on this below).

The learning error is classically viewed as composed from a variance term and a bias term \cite{Breiman1998}. The bias term measures how far the target concept $tc$ is from the classifier search space $\cal H$, that is, from the best classifier $h^*$ in this search space. The variance term measures how far away one can wander from $h^*$, wrongly selecting other classifiers in $\cal H$ (overfitting).

The comparison of the first and second baseline algorithms gives some insight into the intrinsic difficulty of the problem. Stochastic gradient (LMS) will find the global optimum for criterion given by Equation \ref{eq:convex}, but this solution optimizes at best the training error. The comparison between the solutions respectively found by LMS and the simple evolutionary algorithm will thus reflect the learning variance term.

Similarly, the comparison of the first baseline algorithm and AdaBoost gives some insight into how the ensemble improves on the base weak learner; this improvement can be interpreted in terms of variance as well as in terms of bias (since the majority vote of decision stumps allows for describing more complex regions than simple separating hyperplanes alone). 

\subsection{Experimental Setting}

The parameters for the LMS algorithm (see Figure \ref{fig:AlgoLMS}) are as follows: the training rate, set to $\eta(t)=1/(n\sqrt{t})$, decreases over the training epochs; the maximum number of epochs allowed is $T=10000$; the stopping criterion is when the difference in the error rates over two consecutive epochs, is less that some threshold $\epsilon$ ($\epsilon=10^{-7})$. Importantly, LMS requires a preliminary normalization of the dataset, (e.g. $\forall i = 1\ldots n,~{\bf x}_i \in [-1,1]^d$). The final result is the error on the test set, averaged over 10 runs for each fold (because of the stochastic reordering of the training set) and averaged over 10 folds.
\begin{figure}
\caption{Least-mean square training algorithm.}
\label{fig:AlgoLMS}
\begin{center}
\fbox{\parbox{\linewidth}{
\begin{enumerate}
\item Initialize $\mathbf{w}=0$ and $b=0$
\item For $t=1\ldots T$:
\begin{enumerate}
\item Shuffle the dataset ${\cal E} = \{ ({\bf x}_i,y_i),~i = 1\ldots n \}$
\item For $i=1\ldots n$:
\begin{eqnarray*}
a_i        & = & {<{\bf w},{\bf x}_i>}-{b}\\
\Delta_i   & = & 2 \eta(t) (a_i - y_i)\\
\mathbf{w} & = & \mathbf{w} + \Delta_i \mathbf{x}_i\\
b          & = & b - \Delta_i
\end{eqnarray*}
\item $Err_t = \sqrt{\frac{1}{n}\sum_{i=1\ldots n} (a_i - y_i)^2}$ (RMS error)
\item If $|Err_t - Err_{t-1}| < \epsilon$, stop 
\end{enumerate}
\end{enumerate}
}}
\end{center}
\end{figure}

The classical AdaBoost algorithm \cite{Freund1996} uses simple decision stumps \cite{Schapire1998}, and the number of Boosting iterations is limited to $2000$. Decision stumps are simple binary classifiers that classify data according to a threshold value on one of the features of the data set. If the feature value of a given data is less (or greater) than the threshold, the data is assigned to a given class, otherwise it is assigned to another class. Decision stumps are trained deterministically, by looping over all features and all features threshold for a given training dataset, selecting the feature, threshold, and comparison operation on the threshold ($>$ or $<$) that maximize the classification accuracy on the training data set. Decision stumps are the simplest possible linear classifiers, but generate good results in combination with AdaBoost.

The elementary evolutionary algorithm is a real-valued generational GA using SBX crossover, Gaussian mutations, and tournament selection. The search space is $\RR^{d+1}$ for binary classification problems, and  $\RR^{2d+2}$ for ternary classification problem, where $d$ is the number of attributes in the problem domain. The evolutionary parameters are detailed in Table \ref{tab:ParamGA}.
\begin{table}
\caption{Parameters for the real-valued GA.}
\label{tab:ParamGA}
\begin{tabular}{ll} % {p{0.52\linewidth}p{0.38\linewidth}}
Parameter & Value\\
\hline
Population size                      & $500$ \\
Termination criteria                 & $100000$ fitness evaluations\\
Tournament size                      & $2$ \\
Initialization range                 & [-1,1]\\
SBX crossover prob.            & $0.3$\\
SBX crossover $n$-value              & $n=2$\\
Gaussian mutation prob.        & $0.1$\\
Gaussian mutation std. dev. & $\sigma=0.05$
\end{tabular}
\end{table}
% Within \Off\ and \On\, the selection and validation sets are respectively set to 2/3 and 1/3 of the data, randomly drawn. 
All experiments with the real-valued GA rely on the C++ framework Open BEAGLE \cite{Gagne2006,Gagne2006b}.
%%% mettre dans la référence biblio
% \footnote{Available at \url{http://beagle.gel.ulaval.ca}.} 

\section{Results}
\label{sec:EvalMethods}

This section reports on the experimental results obtained by \Off\ and \On, compared to the three baseline methods respectively noted LMS (optimal linear classifier), GA (genetically evolved linear classifier) and Boosting (ensemble of decision stumps), on the six UCI data sets described in Table \ref{tab:DataSetsDescription}. For each method and problem,  the average test error (over 100 independent runs as described in Section \ref{sec:Experiments}) and the associated standard deviation are displayed in Table \ref{tab:Results}. The average computational effort of \Off\ for a run ranges from $30$ seconds (on problem \texttt{bld}) to $20$ minutes (on problem \texttt{spa}), on AMD Athlon 1800+ computers with 1G of memory. For \On, the average computational effort for a run ranges from $2$ hours (on problem \texttt{pid}) to $24$ hours (on problem \texttt{spa}), on the same computers.

\begin{table*}[t]
\caption[Results on the UCI datasets.]{Results on the UCI datasets based on 10-folds cross-validation, using 10 independent runs over each fold. Values are averages (standard deviations) over the 100 runs. Statistical tests are $p$-values of paired $t$-tests on the test error rate compared to that of the best method on the dataset (in bold).}
\label{tab:Results}
\begin{center}
\begin{tabular}{cccccc}
Measure              & LMS                & GA                 & Boosting            & \Off                 & \On\\\hline
\multicolumn{6}{c}{\texttt{bcw}}\\\hline
Train error          & $3.9\%$ ($0.2\%$)  & $1.8\%$ ($0.2\%$)  & $0.0\%$ ($0.0\%$)   & $1.4\%$ ($0.2\%$)    & $0.4\%$ ($0.4\%$) \\
Test error           & $4.0\%$ ($1.6\%$)  & $\mathbf{3.2\%}$ ($1.7\%$)  & $5.3\%$ ($2.0\%$)   & $3.4\%$ ($1.7\%$)    & $3.5\%$ ($2.0\%$)\\
Test error $p$-value & $0.00$             & --                 & $0.00$              & $0.09$               & $0.04$\\
Ensemble size        & --                 & --                 & $291.6$ ($68.2$)    & $235.6$ ($66.8$)     & $116.3$ ($278.2$) \\\hline
\multicolumn{6}{c}{\texttt{bld}}\\\hline
Train error          & $29.8\%$ ($0.9\%$) & $25.4\%$ ($1.2\%$) & $0.0\%$ ($0.0\%$)   & $20.9\%$ ($1.5\%$)   & $18.9\%$ ($2.0\%$)\\
Test error           & $30.4\%$ ($6.6\%$) & $32.7\%$ ($6.6\%$) & $30.4\%$ ($5.4\%$)  & $\mathbf{29.2\%}$ ($7.4\%$)   & $29.5\%$ ($8.4\%$)\\
Test error $p$-value & $0.04$             & $0.00$             & $0.14$              & --                   & $0.64$\\
Ensemble size        & --                 & --                 & $1081.4$ ($166.1$)  & $301.0$ ($37.9$)     & $294.1$ ($154.2$)\\\hline
\multicolumn{6}{c}{\texttt{bos}}\\\hline
Train error          & $32.2\%$ ($1.3\%$) & $23.4\%$ ($4.1\%$) & $0.0\%$ ($0.0\%$)   & $16.7\%$ ($1.9\%$)   & $20.9\%$ ($2.3\%$)\\
Test error           & $34.0\%$ ($6.7\%$) & $30.7\%$ ($7.5\%$) & $26.9\%$ ($4.2\%$)  & $\mathbf{22.7\%}$ ($5.7\%$)   & $26.2\%$ ($7.2\%$)\\
Test error $p$-value & $0.00$             & $0.00$             & $0.00$              & --                   & $0.00$\\
Ensemble size        & --                 & --                 & $761.1$ ($40.8$)    & $303.8$ ($41.4$)     & $2960.9$ ($2109.3$)\\\hline
\multicolumn{6}{c}{\texttt{cmc}}\\\hline
Train error          & $51.6\%$ ($0.4\%$) & $45.7\%$ ($1.4\%$) & $43.3\%$ ($0.7\%$)  & $42.9\%$ ($1.2\%$)   & $43.9\%$ ($1.4\%$)\\
Test error           & $51.8\%$ ($2.5\%$) & $50.4\%$ ($3.9\%$) & $\mathbf{46.8\%}$ ($2.9\%$)  & $\mathbf{46.8\%}$ ($3.9\%$)   & $47.7\%$ ($3.9\%$)\\
Test error $p$-value & $0.00$             & $0.00$             & $0.99$              & --                   & $0.04$\\
Ensemble size        & --                 & --                 & $4000.0$ ($0.0$)    & $326.4$ ($35.7$)     & $2707.7$ ($1696.1$)\\\hline
\multicolumn{6}{c}{\texttt{pid}}\\\hline
Train error          & $22.0\%$ ($0.6\%$) & $20.2\%$ ($0.7\%$) & $0.6\%$ ($0.5\%$)   & $19.8\%$ ($0.7\%$)   & $20.0\%$ ($0.8\%$)\\
Test error           & $\mathbf{22.8\%}$ ($3.5\%$) & $24.2\%$ ($3.9\%$) & $28.1\%$ ($5.0\%$)  & $24.0\%$ ($4.0\%$)   & $24.0\%$ ($3.9\%$)\\
Test error $p$-value & --                 & $0.00$             & $0.00$              & $0.00$               & $0.00$\\
Ensemble size        & --                 & --                 & $1978.1$ ($43.0$)   & $309.5$ ($37.6$)     & $1196.3$ ($765.7$)\\\hline
\multicolumn{6}{c}{\texttt{spa}}\\\hline
Train error          & $11.1\%$ ($0.4\%$) & $7.9\%$ ($0.5\%$)  & $1.4\%$ ($0.1\%$)   & $6.1\%$ ($0.2\%$)    & $7.6\%$ ($0.8\%$)\\
Test error           & $11.3\%$ ($1.2\%$) & $9.0\%$ ($1.3\%$)  & $\mathbf{5.7\%}$ ($0.8\%$) & $6.7\%$ ($1.2\%$) & $8.3\%$ ($1.4\%$)\\
Test error $p$-value & $0.00$             & $0.00$             & --                  & $0.00$               & $0.00$\\
Ensemble size        & --                 & --                 & $2000.0$ ($0.0$)    & $331.1$ ($28.4$)     & $6890.0$ ($2938.1$)
\end{tabular}
\end{center}
\end{table*}

With respect to the baseline algorithms, a first remark is that the LMS-based classifier is significantly outperformed by all other methods, on all problems but one ({\tt pid}). This is explained as the criterion given by Equation \ref{eq:convex} uselessly over-constrains the learning problem, replacing a set of linear inequalities with the minimization of the sum of quadratic terms. Similarly, the single-hypothesis evolutionary learning is dominated by all other methods on all problems but one ({\tt bcw}). Boosting shows its acknowledged efficiency as it is the best algorithm on two out of six problems (\Off\ and Boosting are both best performers for the {\tt cmc} problem). 

\Off\ is the best method for three out of six problems tested. Compared to AdaBoost, it generates ensemble with lower test error rate on four problems, with a tie for the {\tt cmc} problem, and AdaBoost being the best on {\tt spa} problem. In all cases, the number of classifiers is lower, with an average between $235$ and $335$ classifiers for \Off\ compared with more than $750$ on all problems but {\tt bcw} for Boosting. This is understandable given that the ensembles are built with \Off\ starting from a population of $500$ individuals. This raises the question on whether the evolutionary learning accuracy could be improved by considering larger population sizes. But it should not be forgotten that the decision stumps classifier making the AdaBoost ensembles are significantly simpler than the evolved linear discriminants of \Off. No clear conclusion can thus be made on the relative complexity of the ensembles generated by \Off\ compared to Boosting. 

Despite its larger ensemble size, \On\ is dominated by \Off\ on all problems but \texttt{pid}, where both approaches generate identical test error rates. A tentative explanation stems from the nature of the two approaches, with \Off\ having a clear algorithm organized in two stages, classifiers evolution with diversity-enhancing fitness followed by ensemble construction, while \On\ is more complex, with a succession of ensemble construction and classifiers evolution with diversity-enforcing measure taken relatively to the current ensemble. The dynamics of \On\ is hard to understand, but it can be speculated that the iterative construction of the ensemble (without individual removal) is prone to be stuck in local optima. Indeed, the ``construction path'' taken to build the ensemble begins with a selection of some (supposed poor) individuals at the beginning of the evolution. As these individuals cannot be removed from the ensemble, they significantly influence the choice of other individuals, biasing and possibly misleading the whole process.

\section{Discussion and Perspectives}
\label{sec:Conclusion}

This paper has examined the ``Evolutionary Ensemble Learning for Free''
claim, based on the fact that, since Evolutionary Algorithms maintain a population of 
solutions, it comes naturally to use these populations as a pool
for building classifier ensembles.

Two main issues have been studied, respectively concerned with enforcing
the diversity of the population of classifiers, 
and with selecting the classifiers either in the final population or 
along evolution.

The use of a co-evolution-inspired fitness function, generalizing \cite{Liu2000}, 
was found sufficient to generate diverse classifiers. As already noted, 
there is a great similarity between the co-evolution
of programs and fitness cases \cite{Hillis1990} and the Boosting principles \cite{Freund1996};
the common idea is that good classifiers are learned from good examples, while
good examples are generalized by good classifiers. The difference between Boosting and co-evolution is that in Boosting, the 
training examples are not evolved; instead, their weights are updated.
However, the uncontrolled growth of some weights, typically in the case of noisy
examples, actually appears as the Achilles' heel of Boosting compared to 
Bagging. Basically, AdaBoost can be viewed as a dynamic system \cite{Rudin2004}; 
the possible instability or periodicity of this dynamic system has
undesired consequences on the ensemble learning performance. The use of co-evolutionary ideas, even though the set of ensemble does not evolve, seems to increase the stability of the learning process.

The two EEL frameworks investigated in this paper can be considered as promising. \Off\ constructs ensembles with best performances while needing little modifications over a traditional evolutionary algorithm, with a diversity-enhancing fitness and the construction of an ensemble from the final population. But the size of the ensembles generated suggests that bigger population would lead to bigger and possibly better ensembles. For the sake of scalability, this suggests that the ensemble should be gradually constructed along evolution, instead of considering only the final population. This has been explored with \On, with lesser performance comparing to \Off. It is suggested that ensemble construction with \On\ is prone to be stuck in local minima, so some capability of removing individuals can be beneficial, at the risk of inducing an highly dynamic algorithm. Ultimately, the momentum and dynamics of EEL should be controlled by evolution itself, enforcing some trade-off between exploring new regions and preserving efficient optimization. This will be the subject of future researches.

% This work opens two perspectives for further 
% research. On the one hand, larger classifier pools should be considered to 
% construct ensemble classifier. For the sake of scalability, this suggests 
% that the ensemble should be gradually constructed along evolution,
% instead of considering only the final population. On the other hand, 
% as the current ensemble governs the fitness landscape, the ensemble update
% results in a dynamic fitness landscape. It has been suggested that this
% dynamics was one cause for the lesser performance of \On\ compared to 
% \Off. Ultimately, the momentum and dynamics of EEL should be controlled 
% by evolution itself, enforcing some trade-off between exploring new 
% regions and preserving efficient optimization. 

\section*{Acknowledgments}

This work was supported by postdoctoral fellowships from the ERCIM-SARIT (Europe), the Swiss National Science Foundation (Switzerland), and the FQRNT (Qu\'ebec) to C. Gagn\'e. The second and third authors gratefully acknowledge the support of the \texttt{Pascal} Network of Excellence IST-2002-506 778.

\bibliographystyle{abbrv}
\bibliography{eel-gecco}  

\end{document}